\documentclass{article}
\usepackage{spconf,amsmath,graphicx,hyperref}
\usepackage{multicol,multirow,booktabs}
\usepackage{subfigure}

\usepackage{amssymb}
\usepackage{mathtools}
\usepackage{amsthm}

\title{MAD: Multi-Alignment MEG-to-Text Decoding}
\name{
\parbox{0.77\linewidth}{\centering
\textit{Yiqian Yang}$^{1*}$, \textit{Hyejeong Jo} $^{2*}$\thanks{* Equal contribution}, \textit{Yiqun Duan} $^{3*}$, \textit{Qiang Zhang}$^{1}$, \textit{Jinni Zhou}$^{1}$, \textit{Xuming Hu}$^{1\dagger}$, \textit{Won Hee Lee}$^{2\dagger}$, \textit{Renjing Xu}$^{1\dagger}$, \textit{Hui Xiong}$^{1}$\thanks{$^\dagger$ Corresponding authors} 
}}
\address{
The Hong Kong University of Science and Technology (Guangzhou), People's Republic of China$^1$\\
Department of Software Convergence, Kyung Hee University, Republic of Korea$^2$\\
GrapheneX-UTS HAI Centre, Australia Artificial Intelligence Institute, \\University of Technology Sydney, Australia$^3$
}
\begin{document}
%
\maketitle

\begin{abstract}
Deciphering language from brain activity is a crucial task in brain-computer interface (BCI) research. Non-invasive cerebral signaling techniques including electroencephalography (EEG) and magnetoencephalography (MEG) are becoming increasingly popular due to their safety and practicality, avoiding invasive electrode implantation. However, current works under-investigated three points: 1) a predominant focus on EEG with limited exploration of MEG, which provides superior signal quality; 2) poor performance on unseen text, indicating the need for models that can better generalize to diverse linguistic contexts; 3) insufficient integration of information from other modalities, which could potentially constrain our capacity to comprehensively understand the intricate dynamics of brain activity.

This study presents a novel approach for translating MEG signals into text using a speech-decoding framework with multiple alignments. Our method is the first to introduce an end-to-end multi-alignment framework for totally unseen text generation directly from MEG signals. We achieve an impressive BLEU-1 score on the \textit{GWilliams} dataset, significantly outperforming the baseline from 5.49 to 6.86 on the BLEU-1 metric. This improvement demonstrates the advancement of our model towards real-world applications and underscores its potential in advancing BCI research.
\end{abstract}

\section{Introduction}
\label{sec.introduction}

Decoding language from brain activity is a pivotal goal in neurotechnology, promising to restore communication for individuals with severe motor and speech disabilities and to create novel human-machine interfaces. While early successes relied on invasive Electrocorticography (ECoG) signals~\cite{Willett_2023_ecog_speech_neuroprosthesis_rnn_brain2speech2text,Metzger_2023_ecog_hubert_birnn_brain2speech_brain2text_avatar}, the associated medical risks have motivated a shift towards non-invasive techniques like Electroencephalography (EEG) and Magnetoencephalography (MEG).

However, current non-invasive methods face significant hurdles. EEG-to-text models~\cite{wang2022open_aaai_eeg2text, duan2023dewave_brain2text} often exhibit poor generalization to unseen text, rely on teacher-forcing during evaluation, and can suffer from the ``decoder dominated'' problem, where they memorize text distributions rather than learning a true mapping from neural signals~\cite{jo2024are}. Meanwhile, MEG, despite its superior signal quality, has been underutilized. Previous MEG research has focused on decoding limited word classes or short phrases~\cite{ csaky2023interpretable_meg_many_class, ghazaryan2023trials_MEG_decode_written_text}, evaluated performance only on text seen during training~\cite{yang2024decode}, or was limited to classification tasks rather than open-vocabulary sentence generation~\cite{D_fossez_2023_meg_eeg_clip_pretrain_meta_brain2speech}. A critical gap remains in generating novel, complete sentences from MEG signals.

To address these limitations, we propose MAD, an end-to-end framework for open-vocabulary MEG-to-Text translation designed to generalize to \textbf{unseen text}. Our key insight is that aligning brain signals with multi-level representations from a correlated modality (speech) is more effective than relying solely on text-based loss. We employ a speech-decoding architecture that aligns brain features from a Brain Module~\cite{D_fossez_2023_meg_eeg_clip_pretrain_meta_brain2speech} with corresponding speech features extracted by a pre-trained Whisper model~\cite{radford2023robust_whisper_model_originalpaper}. This multi-alignment occurs at three levels: low-level acoustic features (Mel spectrograms), high-level semantic features (encoder hidden states), and text representations.

Our experiments on the public \textit{GWilliams} dataset~\cite{Gwilliams_2023_dataset_meg_208sensors_27persons_56h} show that MAD achieves a BLEU-1 score of 6.86 on entirely unseen text, evaluated without teacher-forcing. This significantly surpasses the previous state-of-the-art performance. Ablation studies reveal that high-level semantic alignment is the most crucial component for successful decoding. Our main contributions are: 1) We present the first end-to-end framework to translate raw MEG waves into open-vocabulary, unseen text without requiring aids like eye-trackers or teacher-forcing. 2) We are the first to systematically investigate various alignment strategies, demonstrating the superiority of aligning with speech over text modality for this task. 3) We validate our model's effectiveness with extensive experiments, setting a new benchmark for non-invasive brain-to-text decoding.
\section{Related Works}
\label{sec:relatedworks}

Brain-to-text decoding has seen remarkable success with invasive methods like ECoG, which can translate speech and imagined handwriting in real-time for open vocabularies~\cite{Willett_2023_ecog_speech_neuroprosthesis_rnn_brain2speech2text}. However, the need for surgical implantation limits their use, motivating safer non-invasive alternatives. These non-invasive approaches, using EEG and MEG, have yet to achieve robust, open-vocabulary generation, often being restricted to classification tasks~\cite{D_fossez_2023_meg_eeg_clip_pretrain_meta_brain2speech, ghazaryan2023trials_MEG_decode_written_text}. While generative EEG models exist~\cite{wang2022open_aaai_eeg2text, duan2023dewave_brain2text}, they have been criticized for poor generalization and a failure to learn a true brain-text mapping~\cite{jo2024are}. Critically, the most relevant end-to-end MEG model shows high performance on seen text but fails on unseen sentences~\cite{yang2024decode}, highlighting a key challenge. Our work directly addresses this generalization problem by introducing a framework that leverages an auxiliary speech modality. By aligning MEG signals with both low-level acoustic and high-level semantic speech features, our model learns a robust representation for decoding entirely new sentences.
\section{Method}
\label{sec.method}

\begin{figure*}[t]
    \centering
    \includegraphics[width=1\linewidth]{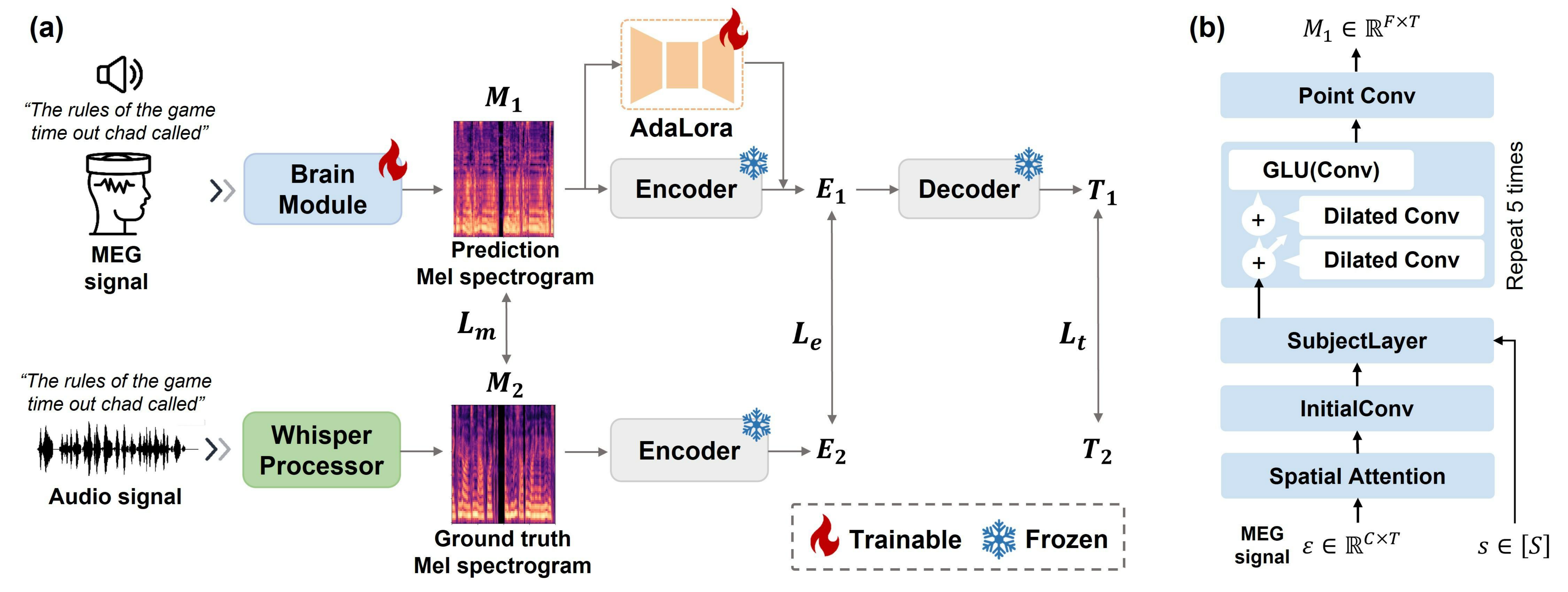}
    \caption{\textbf{(a) Overview of the MAD architecture.} Our model employs a dual-stream design for multi-level alignment between MEG and speech modalities. Alignments are enforced at the level of Mel spectrograms ($M_1, M_2$), encoder hidden states ($E_1, E_2$), and output text ($T_1, T_2$). \textbf{(b) Detailed architecture of the Brain Module} (adapted from~\cite{D_fossez_2023_meg_eeg_clip_pretrain_meta_brain2speech}), which transforms raw MEG signals ($\varepsilon$) into a predicted Mel spectrogram ($M_1$).}
    \label{fig:overview}
\end{figure*}

\subsection{Task and Model Architecture}
Given a raw MEG signal segment $\varepsilon \in \mathbb{R}^{C \times L}$ ($C$ channels, $L$ time points), our goal is to generate the corresponding open-vocabulary text sequence $T$. To achieve this, we introduce MAD, a dual-stream architecture (Fig.~\ref{fig:overview}) that leverages paired speech audio $\Xi$ during training to learn a robust mapping $f: \varepsilon \mapsto T$.

The model's backbone is a pre-trained Whisper encoder-decoder architecture~\cite{radford2023robust_whisper_model_originalpaper}, with its encoder fine-tuned efficiently using AdaLoRA~\cite{zhang2023adaptive_adalora}. The architecture consists of two parallel streams:

\textbf{MEG Stream:} A Brain Module, adopted from~\cite{D_fossez_2023_meg_eeg_clip_pretrain_meta_brain2speech}, first maps the raw MEG signal $\varepsilon$ to a predicted Mel spectrogram $M_1$. This spectrogram is then processed by the Whisper encoder and decoder to produce latent states $E_1$ and the final text output $T_1$.

\textbf{Speech Stream:} The ground-truth audio $\Xi$ is converted to its Mel spectrogram $M_2$, which is then encoded to produce the target latent states $E_2$. These serve as the ground-truth representations for alignment.
    
This design facilitates multi-level alignment by enforcing consistency between the representations derived from MEG and speech.

\subsection{Multi-level Alignment and Loss Functions}
The model is trained end-to-end by optimizing a composite loss function $L$, which is a weighted sum of three components enforcing alignment at different hierarchical levels:
\begin{equation}\label{loss_function}
L = \lambda_m L_m + \lambda_e L_e + \lambda_t L_t
\end{equation}
where $\lambda_m, \lambda_e, \lambda_t$ are balancing hyperparameters.

\textbf{Acoustic-level Alignment ($L_m$).} To align the low-level acoustic features, we employ a symmetric contrastive loss inspired by CLIP~\cite{radford2021learning_clip_origin} between the predicted ($M_1$) and ground-truth ($M_2$) Mel spectrograms. For a batch of $N$ pairs, we compute a scaled cosine similarity matrix $S_{ij} = (\text{norm}(M_{1,i}) \cdot \text{norm}(M_{2,j})^T) / \tau$ with a learnable temperature $\tau$. The loss is the symmetric cross-entropy over this matrix:
\begin{equation}
L_m = \frac{1}{2N} \sum_{i=1}^{N} \left[ \text{CE}(S_{i,:}, i) + \text{CE}(S_{:,i}, i) \right]
\end{equation}

\textbf{Semantic-level Alignment ($L_e$).} To align the high-level semantic representations, we minimize the distributional distance between the encoder hidden states from the MEG stream ($E_1$) and the speech stream ($E_2$) using the Maximum Mean Discrepancy (MMD) loss~\cite{borgwardt2006integrating_mmd_origin}. Given batches of hidden state vectors $X = \{E_{1,i}\}_{i=1}^N$ and $Y = \{E_{2,i}\}_{i=1}^N$, the squared MMD is estimated empirically with a kernel function $k(\cdot, \cdot)$:
\begin{align}
    L_e = \frac{1}{N(N-1)}\sum_{i \neq j} & \left[ k(X_i, X_j) - 2k(X_i, Y_j) + k(Y_i, Y_j) \right]
\end{align}

\textbf{Text-level Supervision ($L_t$).} The final output is supervised using the standard cross-entropy loss. Let $T_2$ be the ground-truth text sequence represented by one-hot vectors $p_{n,j}$ for each token $j$ in each sample $n$ of a batch. Let $T_1$ be the model's predicted probability distribution $\hat{p}_{n,j}$ over the vocabulary $C$. The loss is defined as:
\begin{equation}
L_t = - \frac{1}{NJ} \sum_{n=1}^{N} \sum_{j=1}^{J} \sum_{c=1}^{C} p_{n,j,c} \log(\hat{p}_{n,j,c})
\end{equation}
where $N$ is the batch size and $J$ is the sequence length.

\section{Experiments}
\subsection{Dataset and Preprocessing}
\label{sec.dataset}
We use the \textit{GWilliams} dataset~\cite{Gwilliams_2023_dataset_meg_208sensors_27persons_56h}, which contains MEG recordings from 27 English-speaking participants listening to four distinct stories. To ensure a rigorous evaluation of generalization, we split the data by story: ``cable spool fort'' for testing, ``lw1'' for validation, and the remaining two for training. This guarantees no sentence overlap between the train and test sets. Details of the splits are in Table~\ref{dataset_split_details}.

For preprocessing, raw MEG signals were band-pass filtered between 1-40 Hz and downsampled to 100 Hz. We then extracted 4-second windows with a 1-second stride, applying a random temporal shift of $\pm$0.5 seconds for data augmentation. Corresponding 4-second audio segments were converted to 80-bin Mel spectrograms using the original Whisper configuration~\cite{radford2023robust_whisper_model_originalpaper}.

\begin{table}[!h]
    \vspace{-10pt}
    \centering
    \caption{Details of the story-based data splits. The 'Overlap words' column shows the number (and percentage) of unique words in the test set that also appear in the training set.}
    \resizebox{1.0\linewidth}{!}{
    \begin{tabular}{cccccll}
    \toprule
        Split&  Segments&  Unique sentences&  Words& Unique words &Overlap sentence &Overlap words\\\midrule
        train&  133966&  13266&  150497&  2776& -&-\\
        validation&  14896&  1387&  156027&  478& -&-\\
        test&  31115&  3151&  355654&  805& 0&371(46\%)\\ \bottomrule
    \end{tabular}
    }
    \label{dataset_split_details}
    \vspace{-15pt}
\end{table}

\subsection{Implementation and Evaluation}
\label{subsec.implementation_evaluation}
All models were trained for 5 epochs on a single NVIDIA 4090 GPU using the AdamW optimizer, a learning rate of 3e-4, and a batch size of 32. For our MAD model, loss weights were set to $\lambda_m=1$, $\lambda_e=0.01$, and $\lambda_t=1$. We evaluate performance using BLEU-1~\cite{papineni2002bleu_orig}, ROUGE-1~\cite{lin2004rouge_orig}, BertScore~\cite{zhang2019bertscore_orig}, Character Error Rate (CER)~\cite{martins1991phylogenetic_cer_orig}, and Self-BLEU~\cite{zhu2018texygen_self_bleu_orig} to assess accuracy, semantic similarity, and output diversity.

\subsection{Main Results}
\label{subsec.main_results}
\begin{table}[]
\caption{Comparison with other models. Lo is LoRA, B is brain module. Bert here means Bertscore. Results is obtained without teacher forcing in evaluation. Here, Tr stands for trainable modules. B-1 stands for BLEU-1. R-1 stands for ROUGE-1-F. SB stands for Self-BLEU. RS means randomly selecting sentences from test set as predictions. As we can see, only MAD is much higher than RS on BLEU-1 score. 
\label{compare_with_other_models}
}

\centering
\resizebox{0.99\linewidth}{!}{
\centering
\begin{tabular}{llllccccc}
\toprule
Modality & Method    & Tr & Loss           & \multicolumn{1}{c}{B-1(\%)$\uparrow$} & R-1 (\%)$\uparrow$ & Bert(\%)$\uparrow$ & CER(\%)$\downarrow$  &SB(\%)$\downarrow$ \\ \hline
-      & RS & -                & -           &         5.86                       &      7.20       &    83.73&    87.30     & 96.12\\
MEG      & NeuSpeech~\cite{yang2024decode} & Lo                &$ L_t$           &         5.49                       &      8.43       &    83.98     &    77.02     &99.7\\

MEG      & Wav2vec2CTC~\cite{D_fossez_2023_meg_eeg_clip_pretrain_meta_brain2speech}  & B                 & $L_m$           &          0.55                      &     1.44        &    76.02     &    152.23      &92.67\\
MEG      & MAD      & B              & $L_m+L_e $&                   6.86             &     6.93&     83.39    &  89.82  &85.66\\ 
Noise      & MAD      & B              & $L_m+L_e$ &                   3.87             &     3.16&         83.20&  126.95 &87.54\\ 
MEG      & MAD w/tf      & B              & $L_m+L_e $&                   12.93             &     18.28&   82.87    &  74.31 &83.35\\ 
Noise      & MAD w/tf      & B              & $L_m+L_e $&                   0.19             &     6.68&     59.92    &  87.57 &68.63\\ 
\bottomrule     
\end{tabular}}
\end{table}
We compare MAD against state-of-the-art methods NeuSpeech~\cite{yang2024decode} and Wav2vec2CTC~\cite{D_fossez_2023_meg_eeg_clip_pretrain_meta_brain2speech}, as well as random chance and Gaussian noise inputs to establish performance bounds.

As shown in Table \ref{compare_with_other_models}, our MAD model significantly outperforms all baselines on the primary metric, achieving a \textbf{BLEU-1 score of 6.86}. In contrast, NeuSpeech suffers from extreme repetition (Self-BLEU near 100\%) and a BLEU-1 score lower than random chance, indicating it fails to generalize. The Wav2vec2CTC baseline fails to produce meaningful text. Crucially, MAD maintains a low Self-BLEU score, demonstrating its ability to generate diverse and relevant sentences. A control experiment feeding Gaussian noise into MAD yields a much lower BLEU-1 of 3.87, confirming the model genuinely learns from the MEG signal. With teacher-forcing (MAD w/tf), performance further improves to a BLEU-1 of 12.93, highlighting the model's capacity.

\subsection{Qualitative Analysis}
\label{subsec.qualitative_analysis}
\begin{table}[!h]
    \centering
    \caption{ Transcription results. These are some results obtained without teacher forcing evaluation. \textbf{Bold} for exact matched words, \textit{italy} for similar semantic or pronunciation words. w/ tf means with teacher forcing in evaluation. We lower case results of Wav2vecCTC to give a better visual experience.
    \label{tab:generation_results} }
    
    \resizebox{0.5\textwidth}{!}{
    \begin{tabular}{p{10cm} }
         
     \toprule 
     
     \textbf{Decoding Results on \textit{GWilliams}~\cite{Gwilliams_2023_dataset_meg_208sensors_27persons_56h}} \\ \midrule
     Ground Truth: in one hand and the screwdriver held up high in the other ready to step down into \\
     MAD: As \textbf{to} \textbf{the} worst folk, we are a \textbf{step} \textbf{in} his \textit{floor} \textbf{in} it \textbf{to} separate from prepanded time \\
     MAD w/ tf: of \textbf{one} \textit{otherdriver} \textbf{the} \textbf{to}. \textbf{the} front \textbf{hand} \textbf{to} flip \textbf{up}. \textbf{the} \\
     NeuSpeech: He looked at me \textbf{and} said \textbf{to} me,  \\
     NeuSpeech w/ tf: he \textbf{the} of. \textbf{the} \textbf{other} was \textbf{the}.. \textbf{the} middle.. take on.\\
     Wav2vecCTC: hoas whoistd ban hes hoe leingd s woe stoind hae score mend chroa  \\
     \hline
     Ground Truth: expression and crossed eyes, the tumbleweed in one hand and the \\
     MAD: Primarized. Ribid \textbf{the} fire is \textit{closed}. Your \textbf{eyes} to \textbf{the} thumps \\
     MAD w/ tf: followed \textbf{the} \textbf{eyes} found \textbf{the} of, \textbf{the} other. \textbf{in} \\
     NeuSpeech: He looked at me \textbf{and} said to me, \\
     NeuSpeech w/ tf: heired. \textbf{the} \textbf{the}. he wordsult, of \textbf{the}'s, \textbf{the}\\
     Wav2vecCTC: hien scroucst oin hs oarcsthoins hoer li's b   \\
     \hline
     Ground Truth: the awesomeness of what he intended pulling his eyes \\
     MAD: your \textbf{eyes} panned out your \textbf{eyes} clear \textbf{eye} pain  \\
     MAD w/ tf: \textit{esomeess} \textbf{the} is has to \textbf{the} \textbf{eyes} to  \\
     NeuSpeech: \textbf{He} \textit{looked} at me and said, I'm not sure \textbf{what}'s going on. \\
     NeuSpeech w/ tf: \textbf{he} wayestomeess of \textbf{the} he had to. fingers. his\\
     Wav2vecCTC: is thoane horalaug lind hes schoragthrascre d scrond sfhoanxs s  
    \\\bottomrule
    \end{tabular}
    }
    \vspace{-15pt}
\end{table}

Qualitative results in Table~\ref{tab:generation_results} corroborate MAD's quantitative superiority. It consistently captures semantic content, generating contextually relevant words (e.g., "step", "eyes") and phrases. NeuSpeech, conversely, defaults to generic, repetitive outputs, while Wav2vec2CTC produces incoherent text.

\begin{figure}[t]
    \centering
    \includegraphics[width=1\linewidth]{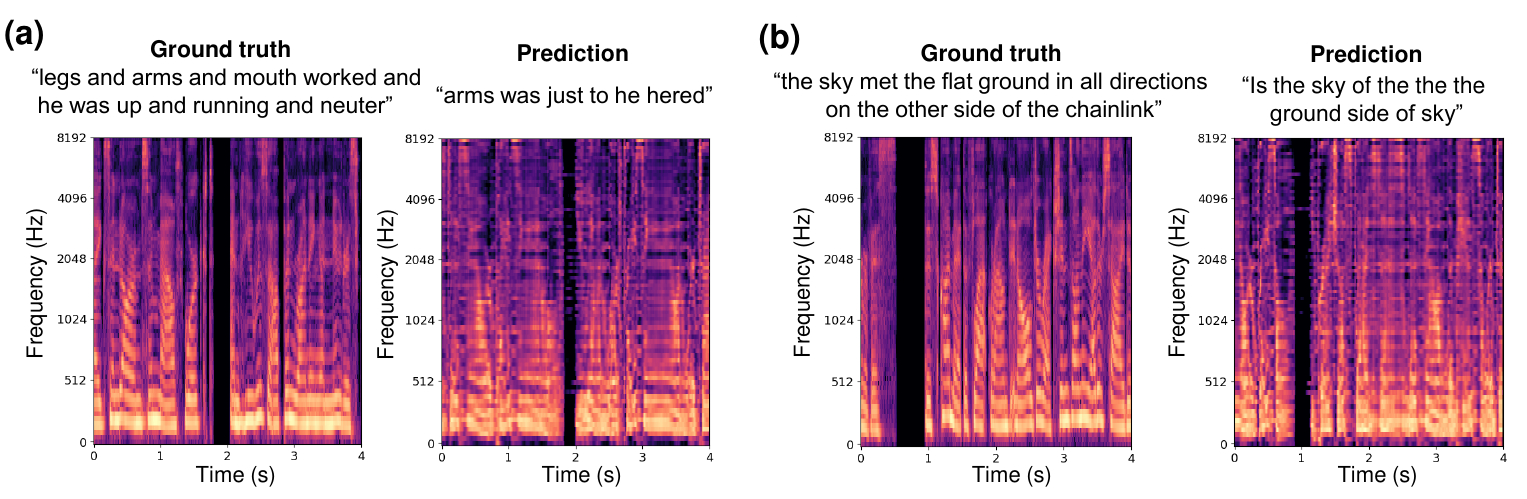}
    \caption{Comparison of ground truth and predicted Mel spectrograms for two test samples. The model captures the overall structure and temporal dynamics (e.g., speech pauses) of the audio.}
    \label{fig:mel_with_text}
    \vspace{-10pt}
\end{figure}

Furthermore, as shown in Figure~\ref{fig:mel_with_text}, the Brain Module successfully predicts the general structure and temporal patterns of the ground-truth Mel spectrograms. While fine-grained details are not perfectly replicated, this confirms that the acoustic-level alignment ($L_m$) is grounded in learning meaningful audio-like features from the brain signal.

\subsection{Ablation Studies}
\label{sec.ablation}
\begin{table}[!h]

\centering
\caption{Performance of the MAD model across different trainable components and loss functions. Where B and Lo denote the brain module and LoRA applied to the encoder, respectively. These results are obtained \textbf{without} teacher forcing in evaluation. Be default, $L_m$ is CLIP loss, $L_e$ is MMD loss, () means loss type replacement. B-1 is the abbreviation of BLEU-1. R-1 is the ROUGE-1-F. SB is self-BLEU. The direction of arrow on metrics indicates better text decoding performance
\label{tab:module_ablation}}
\resizebox{0.8\linewidth}{!}{
\centering
\begin{tabular}{@{}lllccccc@{}}
\toprule
Loss           & Trainables  & B-1 (\%)$\uparrow$ & R-1 (\%)$\uparrow$ & Bert (\%)$\uparrow$ & CER (\%)$\downarrow$ & SB (\%)$\downarrow$ \\ \midrule
$L_m$           & B          &    1.48    &    2.24   &   79.83  &    83.65  &  99.03 \\
$L_e$           & B          &    6.42 &    6.29     &   82.74  &  88.84  &  83.62 \\
$L_e+L_t $     & B          &  4.35 &    4.81    &    84.43     &  80.33  &  95.32 \\
$L_m+L_e(\text{CLIP})$      & B        &    1.22&    1.14&  81.91&    94.85 &  96.16 \\
$L_m(\text{MMD})+L_e$      & B        &    5.44&    5.71&  81.62&    87.95 &  80.55 \\
$L_m+L_e$      & B        &    \textbf{6.86}&    6.93&  83.39&    89.82 &  85.28 \\
$L_m+L_e+L_t$ & B        &     4.29&    4.37&  82.29&  88.40 &  83.95 \\
$L_m+L_e$      & B+Lo        &    0.67   &   0.79     &   81.17  &  87.65 & 99.98 \\
$L_m+L_e+L_t$ & B+Lo        &    6.13&   6.40&  83.14&   91.43 & 99.11\\
 \bottomrule
\end{tabular}}
\end{table}
To understand the contribution of each component, we performed ablation studies summarized in Table~\ref{tab:module_ablation}. The key findings are:

\textbf{High-level semantic alignment ($L_e$) is the most critical component.} Using $L_e$ alone achieves a BLEU-1 score of 6.42, nearly matching the best performance of the full model. This underscores the importance of aligning abstract representations for this task.

    \textbf{Low-level acoustic alignment ($L_m$) is a useful supplement.} While ineffective in isolation (1.48 BLEU-1), adding $L_m$ to $L_e$ provides a marginal but consistent performance boost, increasing the BLEU-1 score to 6.86.

    \textbf{Direct text-level supervision ($L_t$) harms generalization.} Counter-intuitively, including the text reconstruction loss ($L_t$) consistently degrades BLEU-1 scores and causes catastrophic repetition (Self-BLEU $>$ 95\%), likely by preventing the model from learning a robust intermediate representation from the limited data.
    \textbf{Fine-tuning large models is challenging.} Using LoRA to train more parameters leads to severe overfitting (Self-BLEU $>$ 99\%), highlighting the difficulty of adapting large pre-trained models to small, specialized neural datasets without careful regularization.

\section{Conclusion}
\label{sec.conclusion}
In this paper, we presented MAD, a novel end-to-end framework that successfully decodes open-vocabulary, unseen text from raw MEG signals. Our primary contribution is a multi-modal alignment strategy that leverages an auxiliary speech task, demonstrating that aligning brain signals with intermediate representations is more effective than direct brain-to-text mapping for generalization. The key insight from our work is that successful neural decoding hinges on aligning high-level semantic features, whereas direct text-level supervision can paradoxically impair performance in limited-data scenarios. This work provides a new benchmark for non-invasive neural decoding and offers a promising path toward developing practical communication technologies for individuals with severe speech impairments.
\bibliographystyle{IEEEbib}
\bibliography{references}

\end{document}